\documentclass[11pt,a4paper]{article}

\usepackage{fontspec}
\usepackage{polyglossia}
\setdefaultlanguage{english}

\usepackage{amsmath,amssymb}
\usepackage{graphicx}
\usepackage{booktabs}
\usepackage{array}
\usepackage{multirow}
\usepackage{xcolor}
\usepackage[colorlinks=true, linkcolor=blue, citecolor=blue, urlcolor=blue]{hyperref}
\usepackage{cleveref}
\usepackage[round]{natbib}
\usepackage{algorithm}
\usepackage{algpseudocode}
\usepackage{tikz}
\usetikzlibrary{shapes,arrows,positioning,fit,backgrounds}
\usepackage{subcaption}
\usepackage{listings}
\usepackage[normalem]{ulem}
\usepackage{siunitx}
\usepackage{bidi}

\newfontfamily\arabicfont[
    Path=fonts/,
    Extension=.ttf,
    Script=Arabic,
    Scale=1.0,
    Renderer=Harfbuzz
]{Amiri-Regular}

\newfontfamily\bengalifont[
    Path=fonts/,
    Extension=.ttf,
    Script=Bengali,
    Scale=1.0,
    Renderer=Harfbuzz
]{NotoSerifBengali-Regular}

\newfontfamily\chinesefont[
    Path=fonts/,
    Extension=.ttc,
    FontIndex=0,
    Scale=1.0
]{NotoSerifCJK-Regular}

\newfontfamily\cyrillicfont[
    Path=fonts/,
    Extension=.otf,
    Scale=1.0
]{LinLibertine_R}

\newfontfamily\georgianfont[
    Path=fonts/,
    Extension=.ttf,
    Script=Georgian,
    Scale=1.0,
    Renderer=Harfbuzz
]{NotoSerifGeorgian-Regular}

\newfontfamily\greekfont[
    Path=fonts/,
    Extension=.otf,
    Scale=1.0
]{LinLibertine_R}

\newfontfamily\hebrewfont[
    Path=fonts/,
    Extension=.ttf,
    Script=Hebrew,
    Scale=1.0,
    Renderer=Harfbuzz
]{NotoSerifHebrew-Regular}

\newfontfamily\devanagarifont[
    Path=fonts/,
    Extension=.ttf,
    Script=Devanagari,
    Scale=1.0,
    Renderer=Harfbuzz
]{NotoSerifDevanagari-Regular}

\newfontfamily\koreanfont[
    Path=fonts/,
    Extension=.ttc,
    FontIndex=1,
    Scale=1.0
]{NotoSerifCJK-Regular}

\newfontfamily\malayalamfont[
    Path=fonts/,
    Extension=.ttf,
    Script=Malayalam,
    Scale=1.0,
    Renderer=Harfbuzz
]{NotoSerifMalayalam-Regular}

\newfontfamily\tamilfont[
    Path=fonts/,
    Extension=.ttf,
    Script=Tamil,
    Scale=1.0,
    Renderer=Harfbuzz
]{NotoSerifTamil-Regular}

\newfontfamily\telugufont[
    Path=fonts/,
    Extension=.ttf,
    Script=Telugu,
    Scale=1.0,
    Renderer=Harfbuzz
]{NotoSerifTelugu-Regular}

\newfontfamily\thaifont[
    Path=fonts/,
    Extension=.ttf,
    Script=Thai,
    Scale=1.0,
    Renderer=Harfbuzz
]{NotoSerifThai-Regular}

\newcommand{\textar}[1]{{\arabicfont\RL{#1}}}
\newcommand{\textbn}[1]{{\bengalifont #1}}
\newcommand{\textcn}[1]{{\chinesefont #1}}

\newcommand{\embdim}{128}

\title{Symphonym: Universal Phonetic Embeddings for Cross-Script Toponym Matching}

\author{Stephen Gadd\\
\small World Historical Gazetteer, Institute for Spatial History Innovation,\\
\small University of Pittsburgh, USA\\
\small ORCiD: 0000-0003-3060-0181\\
\small \texttt{stephen.gadd@sas.ac.uk}\\
\small \texttt{stephen.gadd@pitt.edu}}

\date{}

\begin{document}

\maketitle

\begin{abstract}
Matching place names across writing systems is a persistent obstacle to the integration of multilingual geographic sources, whether modern gazetteers, medieval itineraries, or colonial-era surveys. Existing approaches depend on language-specific phonetic algorithms or romanisation steps that discard phonetic information, and none generalises across script boundaries. This paper presents Symphonym, a neural embedding system which maps toponyms from thirty-six writing systems into a unified 128-dimensional phonetic space, enabling direct cross-script similarity comparison without language identification or phonetic resources at inference time. A Teacher-Student knowledge distillation architecture first learns from articulatory phonetic features derived from IPA transcriptions, then transfers this knowledge to a character-level Student model. Trained on 73.5 million toponyms drawn from GeoNames, Wikidata, and the Getty Thesaurus of Geographic Names, the Student achieves the highest Recall@1 (89.3\%) and Mean Reciprocal Rank (92.8\%) on the MEHDIE cross-script benchmark (medieval Hebrew and Arabic toponym matches curated by domain experts and entirely independent of the training data), demonstrating cross-temporal generalisation from modern training material to pre-modern sources. An ablation using raw articulatory features alone yields only 45.0\% MRR, confirming the contribution of the neural training curriculum. This revision reports a second model generation, and with it three results that qualify the first: a training-corpus defect caused the original system to learn Chinese characters with Japanese phonetic readings, which we correct and quantify; the encoder was found to weight character \emph{content} far above character \emph{order}, admitting 70.5\% of random anagrams past the production retrieval gate, which targeted negatives reduce to 5.0\% without loss of typographic-error tolerance; and improved retrieval was accompanied by slightly \emph{worse} separation between true and false matches. We report where the method wins decisively (across scripts) and where simple string metrics remain preferable (within the Latin script), and we document an independent out-of-domain deployment on archival personal names.
\end{abstract}


\noindent\textbf{Subject Classifications:} Computational linguistics; Cultural heritage; Digital humanities; Named entity linking; Multilingual information retrieval

\section{Introduction}
\label{sec:introduction}

Place names are the connective tissue of geographic knowledge. A medieval Arabic itinerary, a colonial-era survey, and a modern gazetteer may all refer to the same settlement, but the names they record (rendered in different scripts, shaped by different phonological conventions, and subject to centuries of orthographic drift) share no characters on the page. The same city appears as ``London'', ``Лондон'', ``\textar{لندن}'', and ``\textcn{伦敦}''. Linking such references is prerequisite to the construction of integrated geospatial knowledge bases~\citep{hill2006georeferencing} and to cross-cultural, cross-temporal scholarship, yet no existing computational method bridges script boundaries at scale.

The difficulty is phonetic rather than orthographic. A speaker recognises ``København'' and ``Copenhagen'' as the same city because the sounds are similar, not because the spellings correspond; yet computational toponym matching has relied on string metrics (edit distance, Jaro-Winkler) or on phonetic algorithms designed for individual languages (Soundex for English, Cologne phonetic for German). These fail at script boundaries: no edit distance metric can relate ``\textcn{東京}'' to ``Tokyo''. Historical sources compound the difficulty, introducing pre-standardisation spelling variation (``Deryke/Derico/Diryk'', ``Shotynbaker/Shuttynbaker/Shotyngbaker'') for which no systematic computational bridge exists.

The phonetic encoding of names has a long computational history. Soundex~\citep{odell1918}, Metaphone~\citep{philips1990}, and PHONIX~\citep{gadd1990phonix} encode hand-crafted rules for specific Latin-script languages; combining multiple string metrics improves within-script matching~\citep{recchia2013,santos2018}, but no such ensemble generalises across script boundaries. The cross-lingual embedding literature~\citep{conneau2018,artetxe2018} targets word \emph{meaning} rather than phonetic form, a critical distinction since ``Germany'' and ``Deutschland'' are referentially equivalent but phonetically unrelated, and a phonetic system should not conflate them. Neural methods have been applied to geographic entity matching~\citep{qiu2024geobert,rama2016,mehdie2025}, but existing systems operate within single scripts, require language identification at inference time, or address specific language pairs. \citet{mehdie2025} curated a valuable benchmark of medieval Hebrew-Arabic toponym matches and built a specialist matching system for that pair, but their primary contribution is the benchmark itself rather than a general-purpose cross-script capability.

\emph{Symphonym} addresses this gap. It maps toponyms from thirty-six writing systems into a unified 128-dimensional space in which proximity reflects phonetic similarity. The methodological contribution is a Teacher-Student knowledge distillation architecture~\citep{hinton2015distilling} that grounds the embedding space in universal articulatory phonetics: a Teacher network learns from IPA transcriptions represented as articulatory feature vectors~\citep{mortensen2016}, then transfers this knowledge to a character-level Student that requires no phonetic resources, no language identification, and no grapheme-to-phoneme conversion at inference time.

\paragraph{Contributions of this revision.} This article supersedes an earlier report of the same system~\citep{symphonym2026} and differs from it in four respects, three of which are corrections or qualifications rather than improvements.

\begin{enumerate}
\item \textbf{A second model generation}, trained on 73.5 million toponyms with a vocabulary spanning thirty-six writing systems rather than twenty, raising MEHDIE Recall@1 from 85.2\% to 89.3\%.
\item \textbf{A correction.} The earlier system attributed low CJK-Hiragana similarity to genuine phonological divergence. That explanation was wrong: a defective language tag caused Chinese Han toponyms to be transcribed with \emph{Japanese} readings, so the model learned a similarity structure that was an artefact of grapheme-to-phoneme dispatch (Section~\ref{subsec:contamination}).
\item \textbf{A diagnosis of order insensitivity.} Nothing in the original training made character \emph{order} matter, and 70.5\% of random anagrams of a query cleared the production retrieval threshold. Permutation negatives reduce this to 5.0\% while leaving tolerance of single-character transpositions essentially intact (Section~\ref{subsec:order}).
\item \textbf{An honest account of scope.} Against string baselines across five query types, the method wins overwhelmingly across scripts and \emph{loses} within the Latin script; and the generation that retrieves better separates true from false matches slightly worse (Section~\ref{subsec:tradeoffs}).
\end{enumerate}

\paragraph{Scope.} Symphonym addresses only the \emph{name-matching} component of toponym resolution: it operates purely on phonetic similarity between strings, with no access to geographic coordinates or spatial context. Within a layered retrieval architecture, phonetic similarity functions as a probabilistic prior rather than a final decision mechanism; candidates retrieved via approximate nearest-neighbour search are subsequently filtered by geographic proximity, entity type, and temporal constraints. A similarity score is therefore a statement about a \emph{name}, never about a \emph{place}: ``Newcastle'' in Australia scores exactly as well as the one in England, and any application that accepts matches automatically requires a second, non-phonetic signal. The system is deployed within the World Historical Gazetteer (WHG), where it allows a researcher to search by entering an approximate phonetic rendering in their own script.

\section{Materials and Methods}
\label{sec:methods}

\subsection{Architecture Overview}

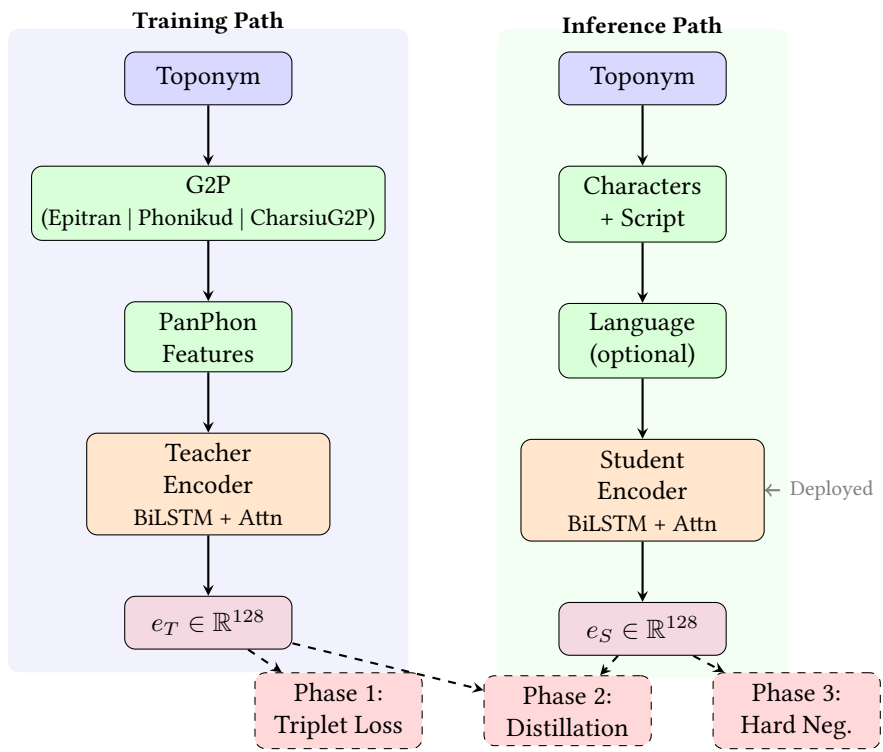
\begin{figure}[ht]
\centering
\begin{tikzpicture}[
    node distance=0.8cm and 1.2cm,
    box/.style={rectangle, draw, rounded corners, minimum width=2.2cm, minimum height=0.7cm, align=center, font=\small},
    widebox/.style={rectangle, draw, rounded corners, minimum width=3.2cm, minimum height=0.7cm, align=center, font=\small},
    input/.style={box, fill=blue!15},
    process/.style={box, fill=green!15},
    encoder/.style={widebox, fill=orange!20, minimum height=1.2cm},
    output/.style={box, fill=purple!15},
    loss/.style={box, fill=red!15, dashed},
    arrow/.style={->, >=stealth, thick},
    dashedarrow/.style={->, >=stealth, thick, dashed},
    label/.style={font=\footnotesize\bfseries},
]

\node[input] (toponym1) {Toponym};
\node[process, below=of toponym1] (epitran) {G2P\\{\footnotesize (Epitran | Phonikud | CharsiuG2P)}};
\node[process, below=of epitran] (panphon) {PanPhon\\Features};
\node[encoder, below=of panphon] (teacher) {Teacher\\Encoder\\{\footnotesize BiLSTM + Attn}};
\node[output, below=of teacher] (temb) {$e_T \in \mathbb{R}^{128}$};

\draw[arrow] (toponym1) -- (epitran);
\draw[arrow] (epitran) -- (panphon);
\draw[arrow] (panphon) -- (teacher);
\draw[arrow] (teacher) -- (temb);

\node[input, right=3.5cm of toponym1] (toponym2) {Toponym};
\node[process, below=of toponym2] (chars) {Characters\\+ Script};
\node[process, below=of chars] (lang) {Language\\(optional)};
\node[encoder, below=of lang] (student) {Student\\Encoder\\{\footnotesize BiLSTM + Attn}};
\node[output, below=of student] (semb) {$e_S \in \mathbb{R}^{128}$};

\draw[arrow] (toponym2) -- (chars);
\draw[arrow] (chars) -- (lang);
\draw[arrow] (lang) -- (student);
\draw[arrow] (student) -- (semb);

\node[loss, below right=0.3cm and -0.5cm of temb] (phase1) {Phase 1:\\Triplet Loss};
\node[loss, right=0.8cm of phase1] (phase2) {Phase 2:\\Distillation};
\node[loss, right=0.8cm of phase2] (phase3) {Phase 3:\\Hard Neg.};

\draw[dashedarrow] (temb) -- (phase1);
\draw[dashedarrow] (temb) -- (phase2);
\draw[dashedarrow] (semb) -- (phase2);
\draw[dashedarrow] (semb) -- (phase3);

\node[label, above=0.1cm of toponym1] {Training Path};
\node[label, above=0.1cm of toponym2] {Inference Path};

\node[font=\scriptsize, text=gray, right=0.2cm of student] (inf) {Deployed};
\draw[->, gray, thick] (inf) -- (student);

\begin{scope}[on background layer]
    \node[fit=(toponym1)(epitran)(panphon)(teacher)(temb),
          fill=blue!5, rounded corners, inner sep=0.3cm] {};
    \node[fit=(toponym2)(chars)(lang)(student)(semb),
          fill=green!5, rounded corners, inner sep=0.3cm] {};
\end{scope}

\end{tikzpicture}
\caption{Teacher-Student architecture. \textbf{Left}: The Teacher converts toponyms to IPA, extracts articulatory features, and encodes to 128-d embeddings (training only). \textbf{Right}: The Student processes raw characters with script/language metadata (deployed at inference). \textbf{Bottom}: Three training phases progressively transfer phonetic knowledge.}
\label{fig:architecture}
\end{figure}

The system employs a Teacher-Student knowledge distillation architecture~\citep{hinton2015distilling} (Figure~\ref{fig:architecture}) in which a Teacher network, trained on articulatory phonetic features, produces target embeddings that a Student network learns to approximate from character sequences alone. At inference time only the Student is required, enabling deployment without phonetic resources. Three principles guide the design: script boundaries are made transparent through deterministic script detection combined with learned script embeddings; embedding similarity reflects phonetic rather than orthographic or semantic similarity, via the Teacher's articulatory feature space; and the deployed model requires no runtime phonetic conversion, language identification, or external resources.
\subsection{Teacher Network}
\label{subsec:teacher}

The Teacher encodes toponyms via IPA transcriptions and articulatory features (Figure~\ref{fig:architecture}, left). PanPhon~\citep{mortensen2016} represents each IPA segment as a 24-dimensional ternary feature vector encoding universal articulatory properties (place, manner, voicing), independent of language identity. The phoneme /b/ thus receives identical features whether appearing in English ``Berlin'', Russian ``Берлин'', or Arabic ``\textar{برلين}''. This grounding enables generalisation: the Teacher learns relationships between articulatory configurations rather than between languages, and that knowledge transfers to any script representing the same sounds. The architecture comprises feature projection, bidirectional LSTM, multi-head self-attention, learned attention pooling, and output projection to \embdim{} dimensions with L2 normalisation.

To obtain fixed-length representations, eight-bin positional pooling divides each IPA sequence into eight bins and averages features within each, yielding $8 \times 24 = 192$ dimensions (PanPhon192). For names of eight segments or fewer each segment occupies its own bin, so the distillation target preserves order at short lengths; for longer names, order information is progressively averaged away.

\subsection{Student Network}
\label{subsec:student}

The Student processes raw character sequences with script and language metadata (Figure~\ref{fig:architecture}, right). Script is detected deterministically from Unicode code points. The vocabulary contains \textbf{114,845 characters} observed across 73.5 million toponyms, thirty-seven script identifiers (thirty-six named writing systems plus an \texttt{OTHER} bucket for unsupported ranges), and 2,438 language codes. Each character maps to a 64-dimensional embedding; script and language contribute 16 dimensions each; and a \textbf{length bucket embedding} (one of sixteen buckets) provides an 8-dimensional conditioning signal broadcast across all positions, yielding a 104-dimensional input per character. The architecture mirrors the Teacher: BiLSTM, self-attention, attention pooling, and output projection with L2 normalisation (8.3M parameters).

\paragraph{Length-Aware Representation.} Toponym lengths vary dramatically, and naive similarity comparison across length disparities produces spurious matches. The length bucket embedding addresses this directly: by conditioning every character representation on a discretised length signal, the model learns to calibrate similarity relative to string length. During training, known languages are randomly replaced with \texttt{<UNK>} at 50\% probability, forcing script-intrinsic learning. Character-level noise augmentation (insertions, deletions, substitutions, and transpositions at 30\% probability) trains robustness to OCR errors and historical spelling variation.

\paragraph{Vocabulary generations are not additive.} The second-generation vocabulary is not the first with entries appended: identifiers are reassigned throughout. A checkpoint paired with the wrong vocabulary raises no error and produces plausible but meaningless embeddings, and the same applies across generations at retrieval time, where a query vector from one generation scored against an index built by another returns confident nonsense. Weights, vocabularies, and index must be replaced together. Published artefacts therefore carry checksums for all three (Section~\ref{subsec:availability}).

\subsection{Training Curriculum}
\label{subsec:training}

Symphonym employs a three-phase curriculum.

\paragraph{Phase 1: Teacher Training.} The Teacher learns to produce embeddings in which phonetically similar toponyms cluster together, using triplet margin loss:
\begin{equation}
\mathcal{L}_{\text{triplet}} = \max\!\big(0,\; \|e_T^a - e_T^p\|_2 - \|e_T^a - e_T^n\|_2 + m\big)
\label{eq:phase1-loss}
\end{equation}
where $a$, $p$, $n$ are anchor, positive, and negative, $m = 0.3$, and $\|\cdot\|_2$ is Euclidean distance. Script-aware negative sampling draws 80\% of negatives from the same writing system as the anchor, forcing fine-grained phonetic discrimination within scripts. Training uses AdamW ($\text{lr} = 10^{-4}$, weight decay $10^{-5}$) with cosine annealing over 50 epochs (final val\_loss 0.0056).

\paragraph{Phase 2: Student-Teacher Alignment.} The Student learns to approximate frozen Teacher embeddings, minimising a combined distillation loss:
\begin{equation}
\mathcal{L}_{\text{distill}} = \alpha \cdot \text{MSE}(e_S, e_T) + \beta \cdot \big(1 - \cos(e_S, e_T)\big)
\label{eq:phase2-loss}
\end{equation}
where $\alpha = \beta = 1.0$. Language dropout and noise augmentation force script-intrinsic learning and input robustness. Training runs 50 epochs, at which point Student-Teacher cosine similarity reached 0.942.

\paragraph{Phase 3: Discriminative Fine-Tuning.} Hard negatives (phonetically similar names from different places) sharpen discrimination, using the same triplet form applied to Student embeddings with $m = 0.3$ and no distillation component retained. Hard negatives are constructed with high orthographic similarity to anchors (same script, same two-character prefix) but no shared place attestations. Training uses AdamW ($\text{lr} = 5 \times 10^{-5}$, batch size 1024) over 30 epochs (final val\_loss 0.0212).

\paragraph{Permutation negatives.} The second generation adds a negative type absent from the first: a reordering of the anchor that preserves its exact character multiset. Section~\ref{subsec:order} gives the motivation and the effect. One constraint is load-bearing rather than a tuning choice. The noise augmentation described above includes adjacent transposition, which teaches the encoder that \texttt{Lodnon} is a typographic error for \texttt{London} and must still match; a permutation negative that happened to be a single adjacent swap would therefore present the identical string as a positive in one place and a negative in another. Candidate permutations are accepted only when at least half of their characters occupy a different index, which keeps the two signals disjoint. Strings too short to reorder, or with fewer than three distinct characters, yield no permutation negative and fall back to a corpus negative.

\subsection{Model Selection}
\label{subsec:selection-model}

Four candidate models were trained under different recipes and scored on one frozen test set under a rule fixed \emph{before} any candidate finished: a candidate is preferred only if recall at the production retrieval gate does not fall \emph{and} separability improves, where separability is the gap between the fifth percentile of retrievable true pairs and the ninety-ninth percentile of false pairs. Table~\ref{tab:arms} reports the outcome.

The selected model leads on collateral measures by margins that are not noise, and is the only candidate with no regression on historical Latin-script pairs; the two candidates leading the headline figure do so by roughly twelve pairs in four thousand.

\paragraph{Training loss does not order candidates across recipes.} One result deserves emphasis because it inverts a common practice. The rejected candidate \textbf{M} achieved the best validation loss at \emph{all three} phases (0.0051, 0.0467, 0.0206), beating the selected model at every one, and was the worst of the four on every downstream measure. Phase loss orders candidates within a fixed batch size and objective; it does not compare across recipes. Any selection procedure for this class of model should score candidates on the retrieval behaviour that will be deployed, not on the curve that training reports.

\subsection{Training Data}
\label{subsec:data}

Training data are extracted from three gazetteer authorities (GeoNames, Wikidata, and the Getty TGN) via a consolidated index of 51.2 million place records, yielding \textbf{73.5 million unique toponyms} after filtering pre-romanised forms and deduplication.

Four curation criteria address data quality. Stratified sampling by script and language pair (capped at 50,000 per bin, with oversampling up to $5\times$ for small bins) prevents domination by high-resource languages. A global vocabulary pass scans the whole corpus. Cross-script positive pairs emerge from density-based clustering (HDBSCAN~\citep{mcinnes2017hdbscan} with $\epsilon = 0.2$) on articulatory feature embeddings, generating pairs only from phonetically coherent groups within each place record. Homonym disambiguation requires that a candidate negative share \emph{no} place attestations with the anchor, preventing one Springfield from serving as a negative for another that may be the same place.

Clustering separates phonetically distinct name families within a place. A Cologne record yields two clusters, Germanic (Köln, Keulen) and Romance (Cologne, Colonia); pairs are generated within clusters but not across them, so the model never learns that phonetically unrelated exonyms are equivalent. A London record with 21 multilingual toponyms yields one dominant cluster of 17 members spanning Arabic, CJK, Cyrillic and Latin scripts (intra-cluster similarity 0.91), while French/Portuguese \emph{Londres}, Bengali \textbn{লন্ডন} and Serbian \emph{Ландон} are correctly isolated.

\subsubsection{IPA Transcription}
\label{subsec:g2p}

IPA transcriptions are generated via three backends: Epitran~\citep{mortensen2018} (extended with 102 new language-script mappings), Phonikud~\citep{kolani2025phonikudhebrewgraphemetophonemeconversion} for Hebrew, and CharsiuG2P~\citep{zhu2022charsiu-g2p} for Chinese topolects and Korean. Dispatch is by detected script \emph{before} language code. Overall IPA coverage is 54.0\%; names without G2P are excluded from Teacher training but are processed by the Student, which generalises from related scripts.

\paragraph{Epitran Extensions.} Epitran ships with grapheme-to-phoneme maps covering perhaps 80--90 language-script pairs, while the corpus contains toponyms in over 1,900 language codes. Accordingly, 102 extension files were developed covering fourteen scripts across diverse families including Austronesian (Acehnese, Balinese, Sundanese), Celtic (Breton, Welsh, Irish), and Turkic (Bashkir, Chuvash, Crimean Tatar). Rules were drafted by prompting multiple commercial large language models in rotation, cross-checking outputs and iterating until convergence. This is best understood as distillation under noise: individual errors propagate as stochastic noise rather than systematic bias, because the Student smooths over Teacher artefacts during distillation and hard-negative training. Scripts relying on extended Epitran nevertheless achieve strong cross-script pass rates despite representing under 1\% of training data each. Intrinsic evaluation would require native-speaker phonological judgements across all 102 languages and was beyond scope; the extensions are validated extrinsically and released with the model.

\paragraph{A systematic failure this treatment does not cover.} The tolerance argument above holds for \emph{stochastic} error. It does not hold when a whole language is dispatched to the wrong backend, which is what occurred for Chinese in the first generation and is analysed in Section~\ref{subsec:contamination}. The distinction matters for anyone adopting this architecture: distillation absorbs noise, and silently learns bias.
\section{Results}
\label{sec:results}

\subsection{MEHDIE Benchmark}
\label{subsec:mehdie}

Symphonym is evaluated on the MEHDIE Hebrew-Arabic historical toponym benchmark~\citep{mehdie2025} using ranking metrics that reflect its design as a candidate generator (Table~\ref{tab:mehdie-ranking}). Four methods are compared: PanPhon192 (raw articulatory features with no neural training), AnyAscii-augmented Levenshtein and Jaro-Winkler string baselines, and Symphonym. PanPhon192 serves as an ablation, using the same G2P pipeline and positional pooling that produce the Teacher's input features but with raw cosine similarity only, thus isolating the contribution of the training curriculum.

\paragraph{Aggregation.} Figures are the unweighted mean across the five testsets, which is the benchmark's own convention and gives an 18-query testset the same weight as a 33-query one. This is stated because the alternative is not innocuous: aggregating the same 137 queries weighted by testset size moves Symphonym's R@1 from 89.3\% to 88.3\% and Levenshtein's from 81.5\% to 81.0\%. To ensure the convention was applied correctly, the string baselines were \emph{re-run} in the same pass rather than quoted from the literature; they reproduce their published values to within 0.1 percentage points on every cell, which is the check that the comparison is like-for-like.

PanPhon192 achieves only 41.1\% R@1 and 45.0\% MRR, roughly half the trained system's performance, and falls well below the string baselines, demonstrating that generic romanisation with edit distance is a considerably stronger cross-script heuristic than raw phonetic features without learned alignment. Symphonym achieves the highest Recall@1 (89.3\%) and MRR (92.8\%), against 85.2\% and 90.8\% for the first generation.

\paragraph{The shape of the improvement.} R@5 is \emph{unchanged} between generations at 97.0\% and R@10 moves 0.6 points, while R@1 gains 4.1 and MRR 2.0. The second generation is therefore not finding answers the first could not find; it is ranking the correct answer first more often. Three independent measurements agree on this reading: the benchmark here, the retrieval evaluation in Section~\ref{subsec:tradeoffs}, and the order-sensitivity bands in Section~\ref{subsec:order}.

\paragraph{Limits of this benchmark.} The benchmark comprises 137 queries across five testsets, so a 4.1-point gain on R@1 corresponds to approximately five or six queries changing rank. The comparison is paired, the same queries scored by both generations, which is stronger than independent samples; it is nevertheless a small benchmark and should not carry more weight than that. Levenshtein retains the highest R@10 (99.4\% against 98.2\%), as it did against the first generation.

\paragraph{Comparison with the MEHDIE System.} Direct comparison with the MEHDIE matching system~\citep{mehdie2025} is not straightforward: their evaluation uses threshold-based F-5 metrics, incommensurable with ranking metrics. Their pipeline, a specialist Hebrew$\leftrightarrow$Arabic transliteration library with Phonetisaurus G2P and PanPhon Hamming distance, is carefully engineered for the phonetic kinship between two Semitic languages. Symphonym addresses a different scope: general-purpose embedding across thirty-six scripts without language-pair-specific resources.

\subsection{Where the Method Wins, and Where It Does Not}
\label{subsec:vstraditional}

A single aggregate score conceals the fact that phonetic embedding and string distance fail in different places. Table~\ref{tab:vstraditional} scores Symphonym against four string metrics over five query types drawn from the same corpus, each query ranked against twenty-one candidates.

The pattern is unambiguous and it is not uniformly favourable.

\begin{itemize}
\item \textbf{Across scripts, string metrics do not function at all.} Symphonym reaches 0.959 R@1 where the best string metric reaches 0.064. This is the capability the system exists to provide, and no amount of tuning closes a gap of that size, because the inputs share no characters.
\item \textbf{Within the Latin script, string metrics are better.} Symphonym's 0.850 R@1 trails \texttt{difflib} ratio (0.909), Jaro-Winkler (0.903) and Levenshtein (0.887). For same-script matching a phonetic embedding is the wrong instrument, and the earlier report's suggestion that same-script variants ``can in any case be handled by traditional edit-distance methods'' understated the case: they are handled \emph{better}.
\item \textbf{On easy historical pairs, string metrics again win} (Levenshtein 0.990 against 0.908); \textbf{on hard historical pairs the methods converge} (Symphonym 0.518, \texttt{difflib} 0.524), with Symphonym ahead of Levenshtein and Jaro-Winkler but not of the best baseline.
\item \textbf{Double Metaphone fails everywhere} (0.05--0.29 R@1), which is the expected result for an algorithm encoding English orthographic conventions applied to a global corpus.
\end{itemize}

The practical implication is that these are complementary channels rather than competitors, and a deployed system should run both. This supports the layered architecture described in Section~\ref{sec:introduction} but sharpens the recommendation: phonetic retrieval should be reserved for the cross-script case and for historical variation, with string metrics carrying same-script refinement.

\subsection{Cross-Script Pair Validation}
\label{subsec:crossscript}

To evaluate cross-script matching comprehensively, 11,723 genuine cross-script pairs were sampled by drawing up to ten from each of 179 script-pair bins. These pairs derive from the same gazetteer sources used for training; this evaluation therefore tests whether trained embeddings produce correct rankings when retrieved over the full index, not whether the model generalises to unseen sources, for which the MEHDIE benchmark serves. Of the 11,723 pairs, 11,535 were scored and 188 had an endpoint no longer present in the current index.

The 0.75 cosine threshold is a high-recall pruning threshold. Testing yields a 90.8\% pass rate, against 90.7\% for the first generation: essentially unchanged, consistent with the finding that the second generation improves ordering rather than reach. Representative script pairs are given in Table~\ref{tab:scriptpairs}. Analysis of sampled embeddings confirms geometric properties appropriate to large-scale retrieval: mean L2 norm 1.000 $\pm$ 0.002, and an effective rank of 11.3 of 128 available dimensions, indicating that the space is neither collapsed nor uniformly spread.

\subsection{A Corrected Result: Chinese Learned Through Japanese}
\label{subsec:contamination}

The earlier report stated that low CJK-Hiragana similarity (mean 0.437) ``reflects genuine phonological divergence rather than embedding instability'', explaining that CharsiuG2P produces Mandarin readings for CJK characters while Epitran produces Japanese readings for Hiragana. \textbf{That explanation was incorrect.}

The mechanism was a defective language tag in the training corpus, under which Chinese Han toponyms were themselves transcribed with Japanese readings. The model therefore learned Chinese characters with Japanese phonetics attached, and the CJK-Hiragana similarity it exhibited was an artefact of grapheme-to-phoneme dispatch rather than a property of the languages. The stated explanation was plausible, cited a real mechanism, and was consistent with the measurement; it was still wrong, because it explained an artefact as a fact.

Correcting the tag and retraining moves CJK-Hiragana mean similarity from 0.437 to \textbf{0.320}. The direction is worth dwelling on: the corrected system scores these pairs \emph{lower}, and lower is right. Chinese Han and Japanese Hiragana genuinely do not sound alike, and the first generation's higher score was the contamination.

The general lesson is stated in Section~\ref{subsec:g2p}: an architecture explicitly designed to tolerate G2P noise will absorb a systematic G2P error just as quietly, and produce a similarity structure that invites a phonological explanation.

\subsection{Character Order Was Barely Learned}
\label{subsec:order}

A recurrent encoder is order-sensitive by construction and cannot be incapable of distinguishing two anagrams. Whether it \emph{has} learned to is a separate question, and one that no evaluation in the earlier report would have detected.

Four bands over 319 queries make it answerable, holding length and character multiset constant so that only order varies: the true variant (a second name attested for the same place, taken from the corpus rather than composed), a single adjacent transposition (a typographic error), a full permutation, a reversal, and an unrelated name. Table~\ref{tab:order} reports the proportion of each band clearing the 0.7 production retrieval gate.

In the first generation, \textbf{70.5\% of random permutations cleared the gate}, against 83.7\% of genuine variants: thirteen points separated a real spelling variant from a meaningless anagram of the query. A true variant outranked its own permutation in only 74.6\% of cases, so roughly one query in four preferred nonsense assembled from its own letters.

The cause was not architectural but curricular. \textbf{Nothing in training ever made order matter.} There were no permutation negatives; and the one order-related signal that did exist, transposition noise augmentation, pushes in the opposite direction, since its purpose is to teach that \texttt{Lodnon} should still match \texttt{London}. A model taught to be invariant to reordering, and never taught that reordering can destroy identity, will weight character content above character order.

Adding permutation negatives under the disjointness constraint of Section~\ref{subsec:training} reduces the permutation band to \textbf{5.0\%} and the reversal band to zero, while the typographic-error band remains at 98.4\% and genuine variants at 85.0\%. A true variant now outranks its own permutation in 95.0\% of cases. The system distinguishes reordering that is a typing error from reordering that destroys the name, which is the distinction required.

\paragraph{The transferable point.} An augmentation introduced to confer robustness teaches invariance, and invariance to the wrong property is indistinguishable from robustness until something measures it. The measurement needs a band rather than an example: a single high cosine between a name and its anagram is uninterpretable, because anagrams share every phoneme and a high score is expected.

\subsection{Retrieval, and a Trade-off}
\label{subsec:tradeoffs}

Both generations were run over one fixed haystack of 1,053,229 names with 4,843 queries, comparing two models rather than two runs (Table~\ref{tab:retrieval}). Every metric improves, but not equally: recall@200 rises 0.0116 while MRR rises 0.0173 and recall@10 rises 0.0283. The second generation is again ordering better rather than reaching further.

\paragraph{A ceiling on re-ranking.} At recall@200 = 0.4908, \textbf{more than half of correct answers are absent from the candidate pool entirely}. Since 200 is the deployed pool size, no downstream re-scoring, re-ranking or filtering can recover them; the ceiling is a property of the embedding space. Reported deficits are concentrated rather than uniform: hard historical pairs recover 4.2\% of correct answers at $k$=200 against 44.8\% for cross-script queries.

\paragraph{Discrimination fell.} Separating true pairs from false ones, measured as AUC, moved from \textbf{0.9324 to 0.9270}. The second generation retrieves and ranks better while separating marginally worse. We report this rather than omitting it because the two properties are genuinely in tension: a space can be flattened to bring more true pairs within reach, at the cost of bringing false ones with them. Any future work optimising for reach should treat separation as a gate rather than a footnote, and any application that thresholds rather than ranks should measure both before adopting a newer generation.

\subsection{Deployment}
\label{subsec:deployment}

Embeddings are quantised to \texttt{int8} and indexed in Elasticsearch with HNSW approximate nearest-neighbour search~\citep{malkov2020hnsw}. Quantisation is not free in principle and measurably free in practice here: scored on the frozen test set, the quantised model reproduces the full-precision model's order-sensitivity band exactly. The production index holds 73,479,069 toponyms at 100\% embedding coverage in 54.1~GB, and query latency averages 15--50ms. Encoding a toponym requires a single forward pass through an 8.3M-parameter network, under 1ms on CPU.

Two practical challenges merit note. High-multiplicity clusters: ``London'' appears in 69 language variants with near-identical embeddings, dominating top-$k$ results, so script-diversity re-ranking or candidate expansion with geographic filtering is necessary. Length sensitivity: long institutional names can produce spurious matches against short toponyms owing to PanPhon192 binning, which the length bucket embedding mitigates without eliminating.
\section{Discussion}
\label{sec:discussion}

The evaluation confirms that a character-level neural encoder can learn phonetically meaningful representations across writing systems without phonetic transcription at inference time. Three aspects merit discussion, two of them cautionary.

\paragraph{Cross-temporal Generalisation.} The most significant positive finding is cross-temporal transfer. The MEHDIE sources, medieval Hebrew and Arabic geographical texts, are entirely independent of the modern gazetteers used for training, and performance on them (89.3\% R@1) indicates general phonetic mappings rather than memorised toponyms. This emerges from the architecture: the Teacher learns phonetic mappings from articulatory features and transfers them through distillation, so the Student inherits a competence grounded in articulatory universals rather than in the orthographic conventions of any period. A model trained on modern authority data thus generalises to historical sources without specialist tuning.

\paragraph{Value of Neural Training.} The PanPhon192 ablation confirms that performance derives from the training curriculum rather than the articulatory features alone: its 45.0\% MRR is half Symphonym's and below even the string baselines. Articulatory features provide necessary phonetic grounding, without which there is no cross-script signal to learn from, but they are radically insufficient alone. The neural architecture learns a non-linear alignment between articulatory feature spaces that distance metrics on raw features cannot capture.

\paragraph{Same-script behaviour is correct, and is not the same as being useful.} Exonyms with genuinely different pronunciations receive appropriately low scores: London/Londres yields 0.474. This is desirable, since a phonetic index \emph{should not} link ``Germany'' to ``Deutschland''; in the production system such links come from a separate places index that groups toponyms attested for the same entity regardless of phonetic similarity. But correct behaviour on exonyms does not make the method the right tool within a script, and Section~\ref{subsec:vstraditional} shows it is not.

\subsection{An Independent Out-of-Domain Deployment}
\label{subsec:customs}

A project on the London customs accounts (1380--1560) applied the first-generation encoder, without fine-tuning, to roughly 48,000 personal-name forms: approximately 38,900 surnames and 5,600 forenames, rather than toponyms. This is informative precisely because it was not conducted by us and not adapted to the domain.

\paragraph{What it shows.} The encoder served as the FAISS candidate generator for spelling-variant retrieval, and in that project's own assessment it was strong for recall but was not used for final ranking, which was left to string-similarity features and human review. That project's considered verdict, that the embedding is ``an excellent recall-oriented candidate generator but a poor precision-oriented discriminator'', matches our own finding in Section~\ref{subsec:vstraditional} from an independent direction and on a different entity type.

\paragraph{What it does not show.} We draw attention to a claim made in the earlier report and withdrawn here. That report described the same project as demonstrating successful \emph{clustering} of merchant-name variants. It does not. Hard clustering was attempted there five separate ways and abandoned in favour of scored candidates with human adjudication, on the grounds that rigid cluster boundaries cannot express uncertainty appropriately for this material; the deployed system leaves the variants cited in that report as separate person records with ranked neighbour lists. Two of the three forename variants quoted were in any case matched by exact lookup against a medieval name dictionary rather than by the model, and the surname variants are Jaro-Winkler 0.93--0.98 of one another, so string matching alone finds them. The example demonstrated less than was claimed for it.

\paragraph{The instructive failure is in the integration, not the model.} That project vendored the tokeniser into its own codebase and introduced a defect in doing so: the vendored tokeniser agreed with the model's own vocabulary on \textbf{0.0000\% of 150,653 character occurrences}, so that \texttt{"william"} reached the encoder as an unrelated sequence of Telugu characters. The defect survived three months undetected, and the reason it did is the point: \emph{the substitution was very nearly a bijection}, so similar names still produced similar vectors, downstream scorers trained on those vectors trained successfully, and every visible signal looked healthy.

What eventually exposed it was not an error but a coefficient. The project's trained surname-pair scorer had learned to \emph{ignore} the embedding, weighting it at +0.045 while string distance did the work. After correction, on identical labels and splits, the embedding cosine became the highest-weighted of twelve features at +1.384, with five-fold cross-validated F1 rising from 0.801 to 0.823 on 1,613 hand-labelled pairs.

\paragraph{Implication for publishing embedding models.} A model whose tokenisation is reimplemented at each integration site will be silently misused, and the failure is invisible to ordinary testing because it degrades quality without producing errors. We therefore distribute a canonical tokeniser with the model, marked as a single block and stamped with a hash of its own source, so that a vendored copy can be checked mechanically against its origin. The mechanism is not hypothetical: the copy in that project has already drifted from the current canonical block, and the check reports it. We recommend the practice generally for published embedding models.

\paragraph{Caveats.} Two limits on this case study should be stated. Everything reported for it concerns the first generation; nothing has been re-run on the second. And there is no precision or recall figure for the deployed name system against a hand-checked sample: the quantitative evidence is the scorer's cross-validated F1 on labelled pairs, not an end-to-end quality measure.

\subsection{Limitations}
\label{subsec:limitations}

\paragraph{Retrieval Ceiling.} More than half of correct answers lie outside the 200-candidate pool (Section~\ref{subsec:tradeoffs}). This is the binding constraint on the deployed system and it cannot be addressed downstream. The deficit is concentrated in hard historical pairs rather than spread evenly.

\paragraph{Discrimination.} The current generation separates true from false pairs slightly worse than its predecessor (AUC 0.9270 against 0.9324) while retrieving better. Applications that threshold rather than rank should evaluate both.

\paragraph{Same-script Matching.} Within the Latin script the method is outperformed by ordinary string metrics (Section~\ref{subsec:vstraditional}). It should not be used alone for same-script matching.

\paragraph{Name Matching Is Not Place Matching.} A similarity score carries no geographic term and cannot separate two places sharing a name. Automatic acceptance requires a second, non-phonetic signal; this is a property of the task, not a deficiency to be trained away.

\paragraph{Training Data Coverage.} Despite stratified sampling, sources retain geographic biases: GeoNames over-represents populated places with official names, Wikidata skews toward encyclopaedic interest, TGN toward art-historical significance. Performance on under-represented scripts and on mundane places lacking multilingual attestations may be weaker.

\paragraph{Tonal Languages.} The architecture does not model tone, which is phonemically contrastive in Chinese, Vietnamese and Thai. PanPhon encodes segmental but not suprasegmental features. In practice tonal minimal pairs are rare in geographic naming.

\paragraph{Confusable Pairs.} Phonetically similar strings necessarily receive high similarity regardless of semantic relationship (Austria/Australia: 0.883; China/Ghana: 0.932). Disambiguation requires evidence beyond phonetic similarity.

\paragraph{G2P Quality.} Individual errors in the 102 extension files are tolerated rather than corrected by the architecture, and performance in languages where the extensions are least accurate may be correspondingly weaker. Section~\ref{subsec:contamination} shows what happens when the error is systematic rather than stochastic.

\section{Conclusion}
\label{sec:conclusion}

Symphonym maps toponyms from thirty-six writing systems into a shared phonetic space using only raw characters at inference time, and achieves the best Recall@1 and MRR of any tested method on an independent benchmark of medieval Hebrew and Arabic material that postdates none of its training data. Across scripts it succeeds where string metrics cannot function at all; within a script it is outperformed by them.

The revision reported here rests as much on what measurement overturned as on what it improved. A published phonological explanation proved to be an artefact of a misrouted grapheme-to-phoneme backend. An encoder assumed to represent character order was found to admit seven anagrams in ten past its own retrieval threshold, because nothing in its training had ever made order matter. A generation that retrieves and ranks better separates true from false slightly worse. None of these was visible in the evaluation originally reported, and each required an evaluation designed to be capable of failing: a band rather than an example, a control that must pass, and a baseline re-run rather than quoted.
\subsection*{Acknowledgments}

This research used the HTC and H2P clusters at the University of Pittsburgh Center for Research Computing and Data (RRID:SCR\_022735), supported by NIH award S10OD028483 and NSF award OAC-2117681 respectively.

The following uses of generative AI tools are disclosed. Claude Sonnet 4.6 (Anthropic), GPT-5 (OpenAI), and Gemini 1.5 Pro (Google) were used to draft grapheme-to-phoneme extension files for the Epitran library (Section~2.5.2), with outputs cross-checked for consistency and validated extrinsically through downstream performance. Claude and GitHub Copilot (Microsoft) were used for coding assistance during system development, and Claude for structural feedback and language editing during manuscript preparation. All content was reviewed, validated, and revised by the author, who takes full responsibility for the accuracy and integrity of the work.

\subsection*{Declaration of Interest Statement}

No relevant financial or non-financial interests to disclose.

\subsection*{Data Availability Statement}
\label{subsec:availability}

All trained models, vocabularies, extension files, and evaluation results for the generation reported here are openly available at \url{https://doi.org/10.5281/zenodo.22767194} \citep{symphonym2026zenodo}; the model, its canonical tokeniser, and an ONNX export are at \url{https://huggingface.co/docuracy/symphonym-v8} \citep{symphonym2026hf}. The artefacts of the first generation remain available and unchanged at \url{https://doi.org/10.5281/zenodo.18682017} and \url{https://huggingface.co/docuracy/symphonym-v7}. Because weights, vocabularies, and index must correspond (Section~\ref{subsec:student}), the published artefacts carry checksums for all three, and the model card states explicitly that embeddings are not comparable across generations. Training code and inference utilities are at \url{https://github.com/WorldHistoricalGazetteer/indexing/tree/main/phonetics}. Training data derive from publicly available sources: GeoNames (\url{https://www.geonames.org/}), Wikidata (\url{https://www.wikidata.org/}), and Getty TGN (\url{https://www.getty.edu/research/tools/vocabularies/tgn/}). The MEHDIE benchmark is available via \citet{mehdie2025}.

\bibliographystyle{plainnat}
\bibliography{references}



\clearpage

\begin{table}[p]
\centering
\small
\begin{tabular}{lrrrr}
\toprule
\textbf{Method} & \textbf{R@1} & \textbf{R@5} & \textbf{R@10} & \textbf{MRR} \\
\midrule
PanPhon192 (ablation)      & 41.1\% & 48.2\% & 52.3\% & 45.0\% \\
Levenshtein + AnyAscii     & 81.5\% & 97.6\% & \textbf{99.4\%} & 88.5\% \\
Jaro-Winkler + AnyAscii    & 78.5\% & 96.3\% & 97.8\% & 86.3\% \\
Symphonym (first generation) & 85.2\% & 97.0\% & 97.6\% & 90.8\% \\
\textbf{Symphonym (this work)} & \textbf{89.3\%} & 97.0\% & 98.2\% & \textbf{92.8\%} \\
\bottomrule
\end{tabular}
\caption{MEHDIE Hebrew-Arabic historical benchmark~\citep{mehdie2025}: unweighted mean across five testsets (137 queries), the benchmark's own convention. String baselines were re-run rather than quoted and reproduce their published values to within 0.1 points, confirming like-for-like aggregation. R@5 is unchanged between generations and R@10 moves 0.6 points, while R@1 gains 4.1: the improvement is in ranking, not reach.}
\label{tab:mehdie-ranking}
\end{table}

\clearpage

\begin{table}[p]
\centering
\small
\begin{tabular}{lrrrrr}
\toprule
\textbf{Query type} & \textbf{Symphonym} & \textbf{Leven.} & \textbf{Jaro-W.} & \textbf{difflib} & \textbf{D.\ Metaphone} \\
\midrule
Cross-script ($n$=4,000)      & \textbf{0.959} & 0.064 & 0.063 & 0.061 & 0.046 \\
Historic, hard ($n$=923)      & 0.518 & 0.402 & 0.420 & \textbf{0.524} & 0.052 \\
Historic, easy ($n$=498)      & 0.908 & \textbf{0.990} & 0.954 & \textbf{0.990} & 0.289 \\
Historic, all Latin ($n$=1,421) & 0.640 & 0.607 & 0.605 & \textbf{0.685} & 0.128 \\
Latin--Latin ($n$=319)        & 0.850 & 0.887 & 0.903 & \textbf{0.909} & 0.201 \\
\bottomrule
\end{tabular}
\caption{Recall@1 against string baselines, each query ranked against twenty-one candidates. Phonetic embedding and string distance fail in different places: across scripts the string metrics do not function at all, while within the Latin script they are the better instrument. The two are complementary channels, not competitors.}
\label{tab:vstraditional}
\end{table}

\clearpage

\begin{table}[p]
\centering
\small
\begin{tabular}{lrr}
\toprule
\textbf{Band (length and character multiset held constant)} & \textbf{First gen.} & \textbf{This work} \\
\midrule
True variant (attested co-referent name)      & 83.7\% & 85.0\% \\
Typographic error (one adjacent transposition) & 100.0\% & 98.4\% \\
\textbf{Permutation (anagram of the query)}   & \textbf{70.5\%} & \textbf{5.0\%} \\
Reversal                                      & 47.6\% & 0.0\% \\
Unrelated name                                & 4.1\% & 2.5\% \\
\midrule
True variant outranks its own permutation     & 74.6\% & 95.0\% \\
\bottomrule
\end{tabular}
\caption{Proportion of each band clearing the 0.7 production retrieval gate, over 319 queries. In the first generation a meaningless anagram was retrievable almost as often as a genuine variant, because nothing in training made character order matter. Permutation negatives close the gap without sacrificing tolerance of typographic errors, which is the distinction required: some reordering is a typing slip, and some destroys the name.}
\label{tab:order}
\end{table}

\clearpage

\begin{table}[p]
\centering
\small
\begin{tabular}{lrrr}
\toprule
\textbf{Metric} & \textbf{First gen.} & \textbf{This work} & \textbf{$\Delta$} \\
\midrule
recall@1    & 0.0593 & 0.0690 & $+$0.0097 \\
recall@10   & 0.2893 & 0.3176 & $+$0.0283 \\
recall@100  & 0.4357 & 0.4547 & $+$0.0190 \\
recall@200  & 0.4792 & 0.4908 & $+$0.0116 \\
MRR         & 0.1383 & 0.1557 & $+$0.0173 \\
\midrule
Discrimination (AUC) & 0.9324 & 0.9270 & $-$0.0054 \\
\bottomrule
\end{tabular}
\caption{Retrieval over one fixed haystack of 1,053,229 names with 4,843 queries, both generations scored on the same corpus and queries. Every retrieval metric improves, but recall@200 least of all: the gain is in ordering. Since 200 is the deployed candidate-pool size, recall@200 = 0.4908 is a ceiling that no downstream re-ranking can raise. Discrimination moved in the opposite direction.}
\label{tab:retrieval}
\end{table}

\clearpage

\begin{table}[p]
\centering
\small
\begin{tabular}{lrrr}
\toprule
\textbf{Script pair} & \textbf{First gen.} & \textbf{This work} & \textbf{$n$} \\
\midrule
Hiragana--Katakana   & 0.981 & 0.993 & 10 \\
Devanagari--Telugu   & 0.976 & 0.975 & 31 \\
Gujarati--Latin      & ---   & 0.957 & 228 \\
Georgian--Latin      & ---   & 0.947 & 210 \\
Devanagari--Latin    & ---   & 0.943 & 681 \\
Cyrillic--Latin      & 0.923 & 0.937 & 1,306 \\
Arabic--Latin        & 0.898 & 0.916 & 786 \\
\midrule
CJK--Hiragana        & 0.437 & \textbf{0.320} & 65 \\
\bottomrule
\end{tabular}
\caption{Mean cosine similarity for representative script pairs. The CJK-Hiragana row is the correction of Section~\ref{subsec:contamination}: the score \emph{falls}, and falling is correct, because the first generation's higher value arose from Chinese toponyms transcribed with Japanese readings. Overall pass rate at the 0.75 threshold is essentially unchanged between generations (90.7\% against 90.8\%).}
\label{tab:scriptpairs}
\end{table}

\clearpage

\begin{table}[p]
\centering
\small
\resizebox{\textwidth}{!}{%
\begin{tabular}{lrrrrr}
\toprule
 & \textbf{First gen.} & \textbf{A} & \textbf{B} & \textbf{G} & \textbf{M} \\
\midrule
Recall at retrieval gate & 87.5 & \textbf{95.2} & 95.5 & 95.3 & 94.2 \\
Separability             & $-$0.2127 & \textbf{$-$0.0574} & $-$0.0510 & $-$0.0593 & $-$0.0758 \\
Permutations clearing gate (\%) & 70.5 & \textbf{5.0} & 6.6 & 6.0 & 4.4 \\
Historic, hard (R@1)     & 0.5179 & \textbf{0.5179} & 0.4995 & 0.5081 & 0.5081 \\
Latin--Latin (R@1)       & 0.8527 & \textbf{0.8495} & 0.8370 & 0.8464 & 0.8370 \\
\midrule
Phase 1 / 2 / 3 val.\ loss & --- & 0.0056 & --- & --- & \emph{0.0051 / 0.0467 / 0.0206} \\
\bottomrule
\end{tabular}}
\caption{Four candidate models scored on one frozen test set under a rule fixed before any candidate finished. Arm \textbf{A} was selected. B and G lead the headline figure by roughly twelve pairs in four thousand, while A leads the collateral measures by larger margins and is the only candidate with no regression on historical Latin pairs. Note arm \textbf{M}: it achieved the best validation loss at all three training phases and is the worst of the four in deployment. Phase loss orders candidates within a fixed objective and does not compare across recipes.}
\label{tab:arms}
\end{table}
\end{document}